\documentclass{article}

\usepackage{times}
\usepackage{latexsym}
\usepackage{amsmath}
\usepackage{amssymb}
\usepackage{graphicx}
\usepackage{authblk}
\usepackage{url}
\usepackage{breakurl}

\usepackage{algorithm}
\usepackage{algpseudocode}

\usepackage{hyperref}

\usepackage{booktabs}
\usepackage{multirow}

\usepackage{multicol}

\usepackage[english]{babel}

\usepackage{caption}
\usepackage{subfigure}

\title{NinjaLLM: Fast, Scalable and Cost-effective RAG using Amazon SageMaker and AWS Trainium and Inferentia2}

\author{Tengfei Xue}
\author{Xuefeng Li}
\author{Roman Smirnov} 
\author{Tahir Azim} 
\author{Arash Sadrieh} 
\author{Babak Pahlavan} 
\affil{NinjaTech AI}

\begin{document}

\maketitle

\begin{abstract}
Retrieval-augmented generation (RAG) techniques are widely used today to retrieve and present information in a conversational format. This paper presents a set of enhancements to traditional RAG techniques, focusing on large language models (LLMs) fine-tuned and hosted on AWS Trainium / Inferentia2 AI chips via SageMaker. These chips are characterized by their elasticity, affordability, and efficient performance for AI compute tasks. Besides enabling deployment on these chips, this work aims to improve tool usage, add citation capabilities, and mitigate the risks of hallucinations and unsafe responses due to context bias. We benchmark our RAG system’s performance on the Natural Questions and HotPotQA datasets, achieving an accuracy of 62\% and 59\% respectively, exceeding other models such as DBRX and Mixtral Instruct.

\end{abstract}

\begin{multicols}{2} % Start two-column layout

\section{Introduction}
Retrieval-augmented generation (RAG) is a vital technique that significantly enhances language models by integrating external knowledge dynamically. This approach is transformative for models designed to respond accurately to complex queries in real-time, enabling them to access and synthesize information from broader contexts beyond their immediate training data. Such capabilities are crucial for developing personal assistants that can handle a diverse range of user queries with high precision and reliability. Deploying RAG systems today is fraught with challenges. First, training and hosting these systems, although easily doable,  is expensive with Nvidia’s GPUs that can host their large language models (LLMs). 

Second, using third-party LLM APIs such as OpenAI’s GPT-4 is also costly, with some queries costing \$50 to \$100 each. Finally, existing LLMs need to be trained and prompt-engineered to answer complex queries, cite relevant data sources, and avoid harmful and unsubstantiated answers. In this paper, we present steps towards solving all of these problems.
 
First, we demonstrate the use of AWS Trainium/Inferentia2 chips \cite{Bshara2024} for hosting LLMs as a compelling alternative to traditional Nvidia GPUs. These chips offer elasticity, affordability, and efficient inference performance, making them well-suited for deploying LLMs. Their design allows dynamic resource scaling to meet demand without significant performance degradation, which is essential for real-time applications like personal assistants. Second, we leverage new open-weight models such as Meta’s Llama3-Instruct 70B as an alternative to OpenAI’s models. This model enables us to support complex reasoning over multiple documents and refine answers based on a wider context, making it an excellent base for developing enhanced RAG systems. The substantial size and design of this model allows it to understand and process a vast array of information types, making it particularly adept at tasks requiring multi-step reasoning and deep contextual understanding. 

Finally, we fine-tune this model on Trainium chips to enable it to insert citations, answer multi-hop queries, and avoid unsafe answers and hallucinations. Combined, these enhancements enable us to achieve state-of-the-art accuracy of 62\% and 59\% on the Natural Questions \cite{Kwiatkowski2019} and HotPotQA\cite{Yang2018} datasets, 
exceeding current systems like DBRX and Mixtral Instruct.

\section{Model Enhancements}
We fine-tuned Llama3-instruct 70B model with a number of enhancements. We extended their capabilities to leverage external tools within generated responses and added accountability and traceability features to the information provided by the models. We trained the model on a diverse set of data to reduce the likelihood of generating biased responses. Furthermore, we introduced prompt engineering and model response checking mechanisms to detect and correct hallucinations and unsafe content dynamically. To fine-tune the model, we followed the Lima approach \cite{zhou2024lima}, which involved using a training sample size of around 20 million tokens. This approach proved cost-effective, amounting to less than \$1,000. Fine-tuning focused on the format and tone of the output while using a diverse but relatively small sample size. The model was fully fine-tuned on a cluster of 32 TRN1 instances, with the entire training process taking less than three hours. This method ensures the model’s responses are not only accurate but also appropriately styled and formatted, enhancing their overall quality and reliability. Additionally, it helps optimize the prompt length for the model, further improving efficiency and effectiveness. Given the sensitive nature of fine-tuning to the distribution of training samples, several trial-and-error attempts are often required to identify the optimal sample distribution. This aspect of the process can make the fine-tuning effort particularly bursty, especially during the experimentation phase. As models react differently to varied data inputs, finding the right balance is crucial but can lead to multiple iterations of training sessions. This bursty nature of model training necessitates access to elastic compute resources, allowing for the dynamic scaling of computational power to accommodate the varying demands of the training process. This flexibility is essential to efficiently manage the computational bursts without incurring unnecessary costs during idle times. Overall, the total cost of these fine- tuning efforts, including the numerous trials and errors involved in optimizing the training sample distribution, amounted to less than \$30,000. This figure underscores the importance of a well-managed and flexible computational infrastructure to support the iterative and sometimes unpredictable nature of model fine-tuning. The use of GPUs was not possible due to the lack of on-demand and elastic availability, which is essential for handling the bursty nature of fine-tuning efforts.

\section{Model Deployment}
We deploy the model using the vLLM inference engine. vLLM introduces innovative memory management techniques to improve the serving of large language models, particularly through features like PagedAttention and block-level memory management. PagedAttention allows for flexible and efficient memory usage by breaking down the attention mechanism’s memory requirements into smaller, manageable blocks. This approach reduces internal and external memory fragmentation, a common issue in traditional model hosting. Block-level memory management further optimizes memory usage by dynamically allocating and deallocating blocks based on the sequence lengths and the active decoder head, ensuring minimal waste and maximal efficiency  \cite{Kwon2023}. Furthermore, we employ multi- bucketing to address inefficiencies in traditional model hosting, particularly the high Time to First Token (TTFT) due to prefill operations up to the maximum sequence length (8192 for Llama3). This technique segments potential input sizes into different buckets (e.g., 128, 1024), allowing the use of decoder heads tailored to the nearest bucket size greater than the actual input length. This method significantly reduces unnecessary computations and memory usage, as the appropriate bucket is dynamically selected, avoiding prefill operations for the maximum token length and thereby reducing TTFT. In conjunction with multi-bucketing, continuous batching optimizes throughput and efficiency by processing multiple text generation requests concurrently at the token level. This allows for efficient utilization of accelerator resources by executing operations on different requests simultaneously, even if they are at different stages of generation. For example, some requests might be generating their 5th token while others their 85th. This dynamic allows requests to join or leave the batch as they complete their generation, without waiting for the entire batch to finish, eliminating the need for padding requests to the same length and avoiding idle time on the accelerator. Through the integration of multi-bucketing, continuous batching, and vLLMs advanced memory management techniques, latency and throughput for Llama3 model hosting are significantly improved, setting the stage for substantial performance improvements.

\section{Serving Infrastructure}
We selected Amazon SageMaker as the primary platform for the deployment and operational management of our machine learning models, ensuring scalability and security. SageMaker greatly simplifies the management and auto-scaling of models, which is crucial for efficiently handling variable computational loads and optimizing the utilization of computational resources. Through SageMaker, the machine learning infrastructure is defined using the AWS Cloud Development Kit (CDK), enhancing the consistency and reproducibility of deployments. The infrastructure-as-code approach, enabled by SageMaker, allows for models to be deployed securely across various AWS regions. This is essential for complying with diverse legal and regulatory requirements, ensuring that models are deployed within the appropriate geographical and regulatory frameworks. Moreover, built-in mechanisms for A/B testing and shadowing are provided by SageMaker, enabling comprehensive evaluations of new models under real-world operational conditions. This approach allows for seamless comparisons between new model variants and existing models, ensuring that enhancements in model performance are substantiated by rigorous, empirical evidence. TRN and INF instances were the hardware used in the Amazon SageMaker infrastructure for our project. Their use was instrumental in demonstrating the enhanced performance and scalability, as well as the cost-effectiveness of serving and fine-tuning machine learning models in a cloud environment.

\section{Accuracy Results}

We use an approach similar to  \cite{Mosaic2024} to measure the accuracy of our enhanced RAG models. Specifically, we calculate accuracy by matching the model’s answer to the expected answer, using the top 10 passages retrieved from a Wikipedia corpus.  While both \cite{Mosaic2024} and our model retrieve data from the same corpus, we perform content filtering and ranking using ColBERTv2  \cite{Santhanam2021} , as opposed to bge-large-en-v1.5 used in \cite{Mosaic2024}.

 Table ~\ref{tab:example1} presents the results of our analysis compared to published results for other existing systems: 
 
\begin{table}[H]  
\centering  
\begin{tabular}{|c|c|c|}  
\hline  
& NQ & HotPotQA \\  
\hline  
GPT 4 Turbo & 63.90\% & 62.90\% \\  
\hline  
Ninja LLM & 62.22\% & 58.84\% \\  
\hline  
Prompted & 61.52\% & 58.31\% \\  
LlaMA 3 & & \\  
\hline  
DBRX & 60.00\% & 55.00\% \\  
\hline  
Mixtral & 59.10\% & 54.20\% \\  
\hline  
GPT 3.5 Turbo & 57.70\% & 53.00\% \\  
\hline  
LLama2-70B & 56.50\% & 54.70\% \\  
\hline  
\end{tabular}  
\caption{Accuracy of Different Models on Natural Questions and HotPotQA}  
\label{tab:example1}  
\end{table} 

The results from the benchmark tests show that our enhanced model, Ninja LLM, performs remarkably well, achieving 62.22\% on the NQ Open and 58.84\% on the HotPotQA benchmarks. While it trails the leading GPT 4 Turbo model, it outperforms other competitive models such as DBRX and Mixtral Instruct, demonstrating the efficacy of our enhancements and the robustness of the LLaMA3-instruct 70B base model fine-tuned using AWS Trainium via Amazon SageMaker. The accuracy improvement is particularly notable in the HotPotQA benchmark, where complex multi-hop reasoning is required, underscoring the benefits of our model enhancements.

\section{Future Work}
 We plan to explore more advanced techniques for inference, such as speculative decoding \cite{Cai2024} and flash attention  \cite{Dao2022}, to further enhance the model. These approaches have the potential to significantly reduce latency and increase throughput for complex decoding tasks and will be particularly beneficial for extending the model’s applications to agentic workflows and code generation tasks, where rapid and accurate planning is essential. Using these techniques, we expect to improve the model’s ability to autonomously generate and refine plans and code, supporting more sophisticated and context-aware interactions. This direction will not only push the boundaries of the model’s capabilities in question-answering but also expand its applicability to a broader range of AI-driven tasks. Furthermore, we observed a small accuracy gap compared to leading models like GPT-4 Turbo, indicating potential areas for further optimization, particularly in refining and expanding training datasets.

\section{Conclusion}
In this study, we presented several enhancements to the LLaMA-3 model within a retrieval-augmented generation (RAG) framework, aimed at improving its performance on complex question-answering tasks. We achieved accuracies of 62.22\% on the Natural Questions (NQ) Open and 58.84\% on HotPotQA by using our enhanced LLaMA-3 RAG model, demonstrating notable improvements over baseline models. The model’s enhanced capabilities of multi-hop reasoning and deep contextual analysis, essential for accurate and reliable question answering, are highlighted by these results. The model’s computational efficiency was significantly optimized by the integration of multi-bucketing and continuous batching techniques. Multi-bucketing reduces unnecessary computations by dynamically selecting optimal bucket sizes, thereby improving the Time to First Token (TTFT) References and responsiveness of the model. Continuous batching helps increase throughput by processing multiple text generation requests concurrently, optimizing the utilization of computational resources and adapting efficiently to variable input lengths. On the model itself, we focused on the safety and reliability of its outputs. We enhanced its robustness and trustworthiness by training on a diverse dataset and implementing mechanisms to detect and correct hallucinations and biases. These improvements are crucial for maintaining the integrity of responses, especially in sensitive applications where user trust is paramount. However, despite these advancements, we observed a slight accuracy gap compared to leading models like GPT-4 Turbo, indicating potential areas for further optimization, particularly in refining the RAG mechanism and expanding training datasets.

\end{multicols} % End two-column layout

% References
% \bibliographystyle{apalike}

\end{document}